\documentclass[journal]{IEEEtran}
\usepackage{multirow}

\ifCLASSINFOpdf
   \usepackage[pdftex]{graphicx}
\else
\fi
\begin{document}
%
\title{flexgrid2vec: Learning Efficient Visual Representations Vectors}
%
%
%

\author{Ali~Hamdi,
        Flora~Salim,
        and~Du Yong~Kim,
\thanks{Ali Hamdi and Flora Salim are with the School of Computing Technologies, RMIT University, Australia, ali.ali@rmit.edu.au, flora.salim@rmit.edu.au.}
\thanks{Du Yong Kim is with the School of Engineering, RMIT University, Australia, duyong.kim@rmit.edu.au.}
\thanks{Manuscript received September 27, 2021; revised 15 January 2022.}}

%
%

\markboth{Ali Hamdi \MakeLowercase{\textit{et al.}}: flexgrid2vec: Learning Efficient Visual Representations Vectors}%
{Ali Hamdi \MakeLowercase{\textit{et al.}}: flexgrid2vec: Learning Efficient Visual Representations Vectors}
%



\maketitle

\begin{abstract}
We propose \textit{flexgrid2vec}, a novel approach for image representation learning. Existing visual representation methods suffer from several issues, including the need for highly intensive computation, the risk of losing in-depth structural information and the specificity of the method to certain shapes or objects. \textit{flexgrid2vec} converts an image to a low-dimensional feature vector. We represent each image with a graph of flexible, unique node locations and edge distances. \textit{flexgrid2vec} is a multi-channel GCN that learns features of the most representative image patches. We have investigated both spectral and non-spectral implementations of the GCN node-embedding. Specifically, we have implemented \textit{flexgrid2vec} based on different node-aggregation methods, such as vector summation, concatenation and normalisation with eigenvector centrality. We compare the performance of \textit{flexgrid2vec} with a set of state-of-the-art visual representation learning models on binary and multi-class image classification tasks. Although we utilise imbalanced, low-size and low-resolution datasets, \textit{flexgrid2vec} shows stable and outstanding results against well-known base classifiers. \textit{flexgrid2vec} achieves $96.23\%$ on CIFAR-10, $83.05\%$ on CIFAR-100, $94.50\%$ on STL-10, $98.8\%$ on ASIRRA and $89.69\%$ on the COCO dataset.
\end{abstract}


%
\IEEEpeerreviewmaketitle

\section{Introduction}
%
%
%
%
\IEEEPARstart{R}{Representation} learning is a fundamental problem in computer vision. Convolutional Neural Networks (CNNs) have been used to learn image features in multiple visual applications. CNNs learn representative image convolutions through receptive fields to capture context information. However, CNNs are limited to local regions and affected by their isotropic mechanism \cite{luo2016understanding}. These challenges make CNNs unable to cope directly with useful structural information of an image or to deal with diverse backgrounds and unrepresentative regions. Therefore, we describe the best solution to learn the convolutions only for the useful image segments (patches) while preserving the overall structural information. To this end, the image may be represented in a graph that has its nodes represent image patches or regions. However, CNNs cannot directly deal with such a graph of irregular structures.

\begin{figure}[!ht] 
\begin{center}
  \includegraphics[angle=0, origin=c, width=0.99\linewidth]{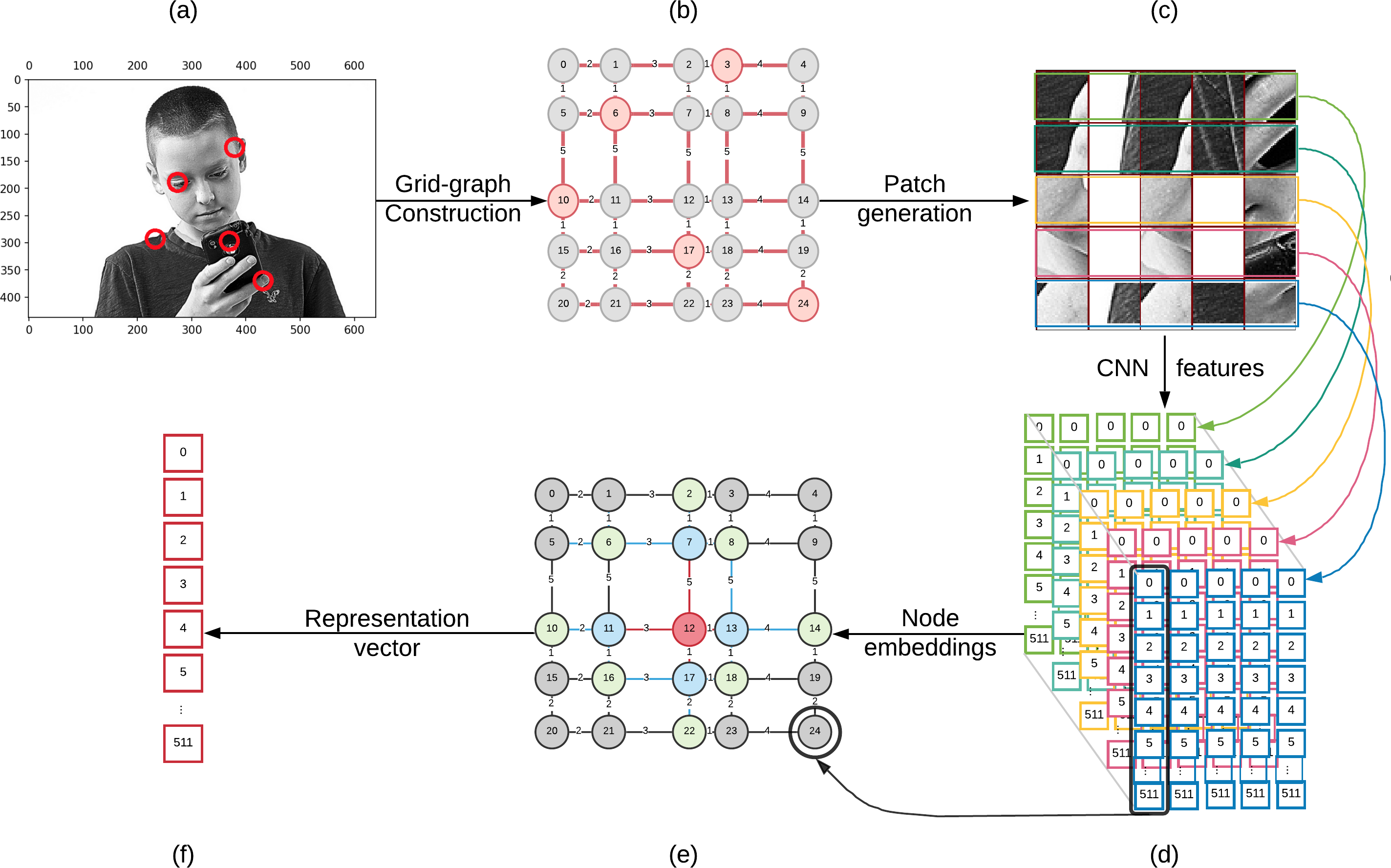}
\end{center}
  \caption{\textit{flexgrid2vec} consists of multiple components: 
  (a) Extracts the image key-points. 
  (b) Visualises the node projection and the constructed grid-graph. 
  (c) Generates patches around node key-points. 
  (d) Computes the convolution features.
  (e) Computes node embeddings learnt with neighbourhood features.
  (f) Produces the final features vector.
  }
\label{GG}
\end{figure}

Multiple techniques have tried to represent an image as a graph to harness the power of representation learning, such as part-based \cite{gao2019graph}, region-based \cite{li2018beyond}, pre-defined skeleton \cite{yan2018spatial} and patch-based grid-graph \cite{zhou2018graph}. Recently, Graph Convolutional Neural Networks (GCN) has been utilised to implement deep convolution networks over graphs. These approaches outperform CNN in encoding long relations between image regions \cite{li2018beyond}. However, these approaches suffer from several issues, such as losing significant local structures and the need for high computation resources. Graph models were developed for visual tasks with multiple objects in an image, such as object detection \cite{Xu_2019_CVPR}, tracking \cite{gao2019graph}, action recognition \cite{Si_2019_CVPR}, person search \cite{Yan_2019_CVPR}, face clustering \cite{Wang_2019_CVPR}, human pose estimation \cite{Zhao_2019_CVPR}, multi-label image recognition \cite{Chen_2019_CVPR} and segmentation \cite{Wang_2019_CVPR}. Constructing a graph in these applications is simple as the semantic components are known in labelled images. However, there is difficulty in dealing with the whole image with no pre-defined semantic labels. The proposed \textit{flexgrid2vec} solves this issue by handling the image as a grid-graph, combined with the power of GCN for visual image classification.

We propose \textit{flexgrid2vec}, which constructs graph nodes only of informative image patches. We locate each graph node on a grid with original image coordinates. Our proposed design maintains the advantages of the global structure of the 2D grid, the local structure of the superpixels or region-based graphs and the flexible structure of the skeleton-based methods. Fig. \ref{GG} shows the main components of the proposed \textit{flexgrid2vec}. It converts an image to a feature vector through a 2D grid-graph that aggregates the node embeddings using multiple CNNs. \textit{flexgrid2vec} incorporates a set of novel contributions, as follows:
\begin{itemize}
  \item A visual representation learning approach provides compact image representations considering significant image patches and preserving their spatial relations on the grid. 
    \item An effective method for learning node embeddings using GCN learnt with neighbourhood features.
    \item A comparative study on multiple image classification tasks with well-known baseline architectures and embedding methods on five benchmark datasets, namely CIFAR-10, CIFAR-100, STL-10, ASIRRA and COCO.
\end{itemize}

The rest of this paper is organised as follows. The related works are discussed in Section \ref{relate_works}. Section \ref{proposal} explains the components of the proposed flexgrid2vec. Sections \ref{results} and \ref{conclusion} show the experiment results and work conclusions, respectively.

\section{Related Work}\label{relate_works}
We discuss recent developments of CNN and GNN architectures for image representation. We then review related work in graph representation to highlight the significance of our proposed method in terms of flexibility and expressiveness. 

\subsection{Convolutional Neural Networks.}
Recently, many CNN architectures have been widely used to learn image representations, such as VGG \cite{simonyan2014very}, ResNets \cite{he2016deep,huang2016deep,he2016identity}, DenseNet \cite{huang2017densely}, MobileNet \cite{Sandler_2018_CVPR,howard2017mobilenets}, and NASNet \cite{zoph2018learning}. The performance of these models depends on the training data, network architecture, and loss function. These models are trained on large scale image datasets, e.g., ImageNet. CNNs employ different data augmentation techniques to achieve high accuracy and solve overfitting issues \cite{masi2016we}. CNN models are being developed to have deeper structures to tackle the increasing size and complexity of training data. The CNN loss functions are being developed to compute effective gradients to learn the most discriminative convolution features. A large body of research in computer vision uses the CNNs, as mentioned earlier, as base models to achieve different visual representations \cite{zhang2019aet,kolesnikov2019revisiting,hamdi2020drotrack} tasks. Recently, these CNNs have advanced image classification accuracy. For example, \cite{he2016deep} achieved high accuracy in different image recognition tasks such as localisation and detection using ImageNet and COCO datasets. However, CNNs neglect useful structures due to the limitations of their receptive fields and isotropic mechanism \cite{luo2016understanding}. 

\subsection{Graph Neural Networks.}
GCN aims to generalise the CNN to accommodate the non-euclidean graph data. GCN includes spectral and non-spectral representations. Spectral representations depend on graph polynomial, eigenvalues, and eigenvectors \cite{kolykhalova2020automated,morisi2016hierarchical}. This process requires extreme computations and results in structure-dependent representations \cite{zhou2018graph}. Non-spectral GCN approaches operate directly on graph spatial neighbours. They adapt with vary-sized neighbourhoods to preserve the advantage of local invariance of the CNN \cite{duvenaud2015convolutional}. GCN models have utilised in different computer vision tasks such as spatiotemporal graph for video analyses \cite{wang2018videos,gao2018watch}, action recognition \cite{yan2018spatial,gao2019changsheng}, and person re-identification \cite{shen2018person}. GCN is combined with Recurrent Neural Networks on cyclic variations of directed graphs to model moving objects \cite{fan2017sanet}. GCN has been combined with Haar transforms \cite{li2020fast,wang2020haar}, attention mechanism \cite{velivckovic2017graph}, Bayesian networks \cite{atzeni2021modeling}, adversarial auto-encoders \cite{pan2018adversarially}. GCN is also utilised with Siamese networks for object tracking \cite{gao2019graph,cui2018spectral,BertinettoVHVT2016Fully}. However, to our knowledge, no GCN work considers the whole image representation. This paper proposes to encode the 2D image into a feature vector based on flexibly constructed grid-graphs.

\subsection{Graph Representation Methods.}
The main contribution of this paper is a new image representation learning method. This idea extends many previous graph construction techniques, such as part-based \cite{gao2019graph}, region-based \cite{li2018beyond}, pre-defined skeleton \cite{yan2018spatial}, and patch-based grid-graph \cite{zhou2018graph}. Table \ref{tbl:graph_constructions} lists descriptions, applications, and limitations of these different graph designs. The part-based approach divides the image into a set of segments or patches \cite{gao2019graph}. These segments are used to construct graph nodes and edges. This approach depends on the spectral graph representation and node embedding to perform visual recognition. However, it loses significant local contextual structures. Region-based, not only with graphs, approaches use image regions for contextual reasoning and perception. They cluster similar image pixels into coherent regions. Graph structures are utilised to model various contexts for image recognition. Region-based approaches for graph representation outperform CNN in encoding long relations between image regions \cite{li2018beyond}. Pre-defined skeleton-based methods are limited to certain shapes such as the human body based on annotated joints or coordinates \cite{yan2018spatial}. Another solution is to construct a grid-graph using non-overlapped patches \cite{zhou2018graph}. The latter approach aims to reduce the number of graph nodes resulting in lower computation cost. However, this approach lacks a deep knowledge of the local structures. We propose \textit{flexgrid2vec} that takes the advantages of the global structure of the 2D grid-graphs, local structure of the superpixels or region-based graphs, and flexible structure of the skeleton-based methods.

\begin{table*}[]
\label{tbl:graph_constructions}
\caption{A comparison between the existing graph construction methods.}
\small
\centering
\begin{tabular}{|p{1.5cm}|p{3cm}|p{3cm}|p{2cm}|p{1.5cm}|}
\hline
Graph & Description & Applications & Limitations & Design \\ \hline
Part-based or Patch-based graph \cite{gao2019graph} & Segment the image into several parts or patches, each of which is represented by a node in the graph & Object tracking. & Loos informative local structures. 
& 
\vspace{.01\linewidth}\centerline{\includegraphics[width=.95\linewidth]{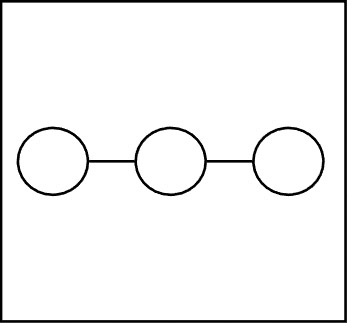}}

\\ \hline
Region-based or Superpixels graph \cite{li2018beyond,cheng2016hfs} & Group similar pixels into a set of groups (superpixels) and construct graph nodes for each group. & Object detection. Saliency detection. Segmentation. Semantic object parsing. & Loos informative local structures. 
& 
\vspace{.01\linewidth}\centerline{\includegraphics[width=.95\linewidth]{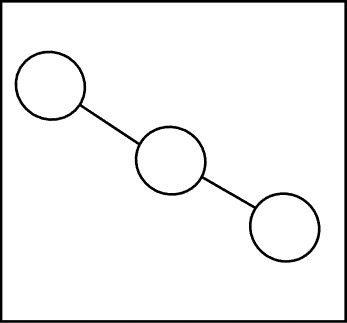}}
\\ \hline
Tree graph \cite{wang2008normalized,wang2014regularized}& An image is represented as a tree graph. A root node refers to the image. Child nodes represent the major components of the images. Leaf nodes are the image pixels. 
& Face recognition. Image segmentation.
& Hard coded.  High-cost  computation. 
& 
\vspace{.01\linewidth}\centerline{\includegraphics[width=.95\linewidth]{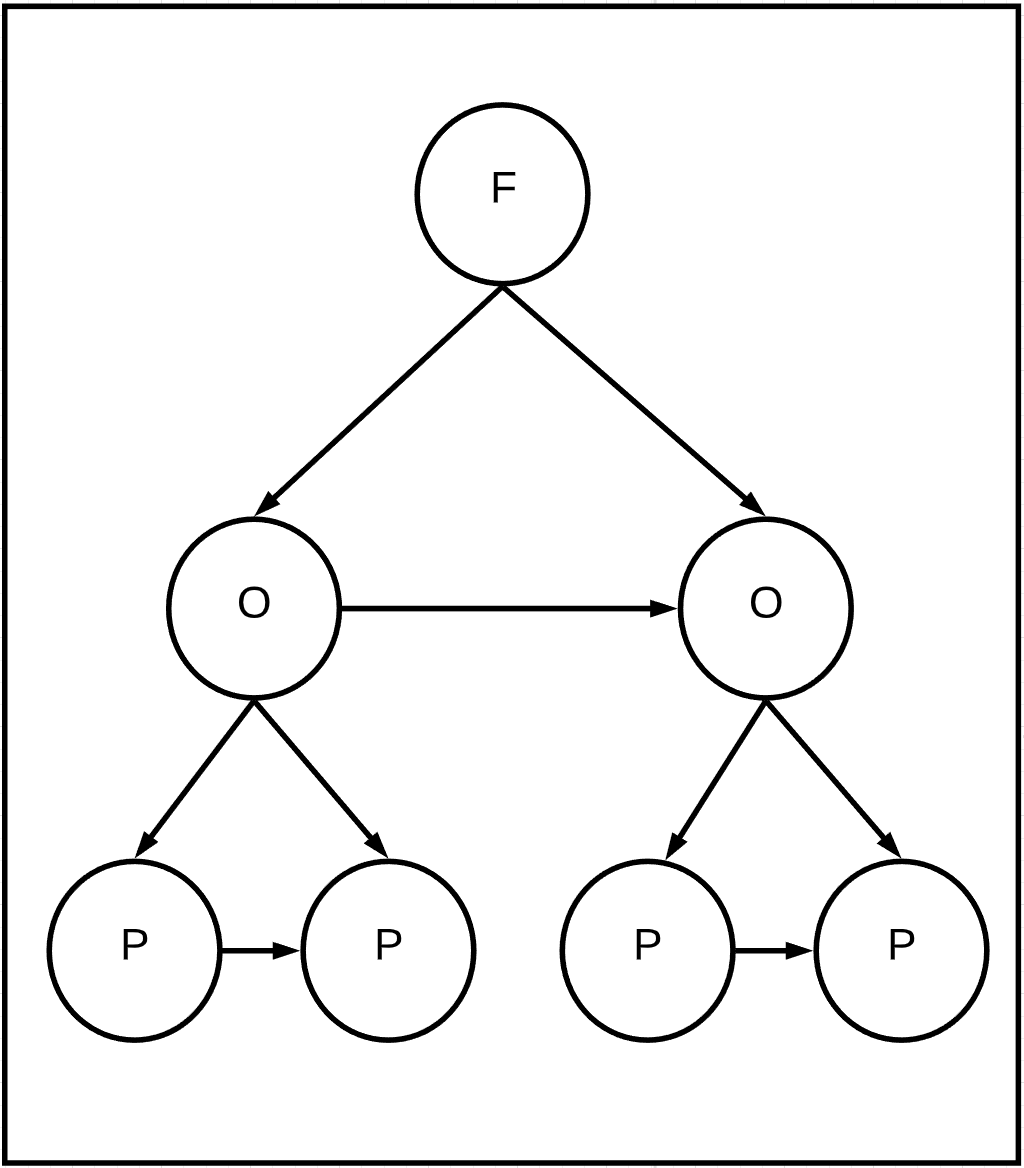}}
\\ \hline
Skeleton \cite{yan2018spatial} & Search for a skeleton point such as the human body. & Human body recognition. & Hard coded.  Shape-dependent. 
& 
\vspace{.01\linewidth}\centerline{\includegraphics[width=.95\linewidth]{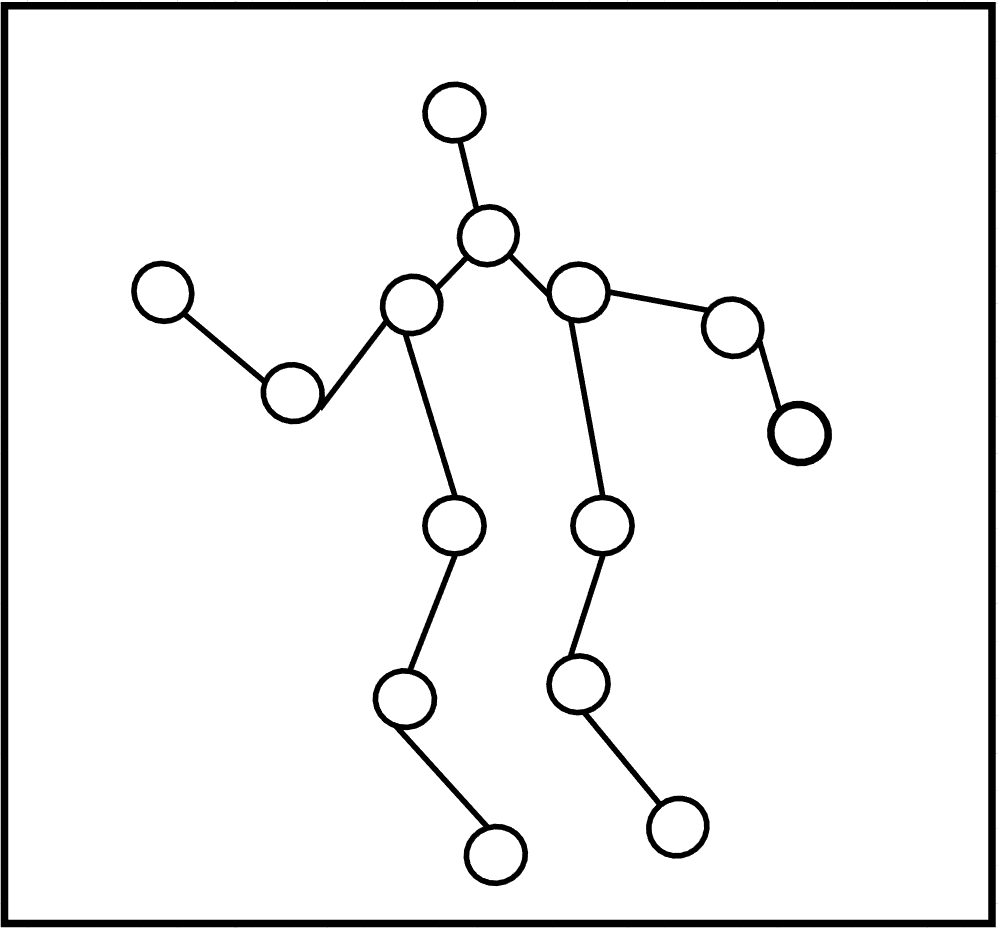}}
\\ \hline
Pixel-level grid-graph \cite{chen2019graph} & Construct a graph node for each pixel in the image. & Image classification. & High-cost computing. 
& 
\vspace{.01\linewidth}\centerline{\includegraphics[width=.95\linewidth]{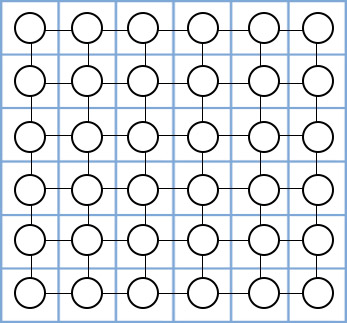}}
\\ \hline

Patch grid-graph \cite{zhou2018graph} & Construct a 2D grid-graph node for patches of non-overlapped pixels in the image. & Image classification. Person re-identification. & High-cost computing. 
& 
\vspace{.01\linewidth}\centerline{\includegraphics[width=.95\linewidth]{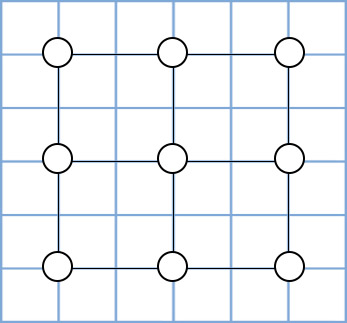}}
\\ \hline

flexgrid2vec (Ours) & Detect representative features, e.g., strong edges or key-points, and build the graph nodes using these points. & Image classification. & Dependent on the quality of key-point detectors. & \vspace{.01\linewidth}\centerline{\includegraphics[width=.95\linewidth]{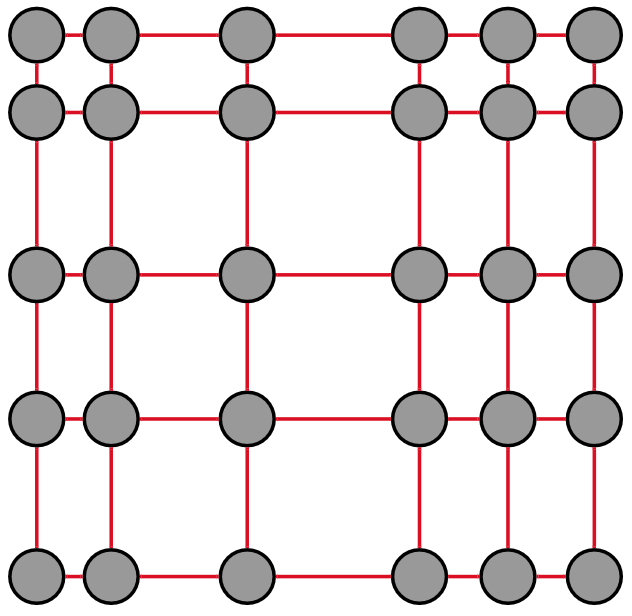}}
 \\ \hline
\end{tabular}
\end{table*}

\begin{table}[!h]
\centering
\caption{Used notations.}\label{notations}
\small
\begin{tabular}{|p{1cm}|p{2.4cm}|p{1cm}|p{2.4cm}|}

\hline
Notation & Description & Notation & Description \\ \hline
$G$ & Graph & $V$ & The set of nodes. \\ \hline
$v$ & A node $v \in V$ . & $E$ & The set of edges in a graph. \\ \hline
$e_{ij}$ & An edge $e_{ij} \in E$. & $N(v)$ & The neighbours of a node $v$. \\ \hline
$u \in N(v)$ & A neighbour of $v$. & $z \in N(u)$ & A 2-step neighbour of $v$, neighbour of neighbour. \\ \hline
$n$ & The number of nodes, $n = |V|$. & $m$ & The number of edges, $m = |E|$. \\ \hline
$d$ & The dimension of a node feature vector. & $X \in R^{n\times d}$ & The feature matrix of a graph. \\ \hline
$x_v \in R^d$ & The feature vector of the node $v$. & $b$ & The dimension of a hidden node feature vector. \\ \hline
$H \in R^{n\times b}$ & The node hidden feature matrix. & $h_v \in R^b$ & The hidden feature vector of node v. \\ \hline
$c$ & The dimension of an edge feature vector. & $X^e \in R^{m\times c}$ & The edge feature matrix of a graph. \\ \hline
\large y & The actual label. & $\large \hat y$ & The predicted label. \\ \hline
\end{tabular}
\end{table}

\section{Overview}\label{problem}

\begin{figure}[!ht] 
\begin{center}
  \includegraphics[angle=0, origin=c, width=0.99\linewidth]{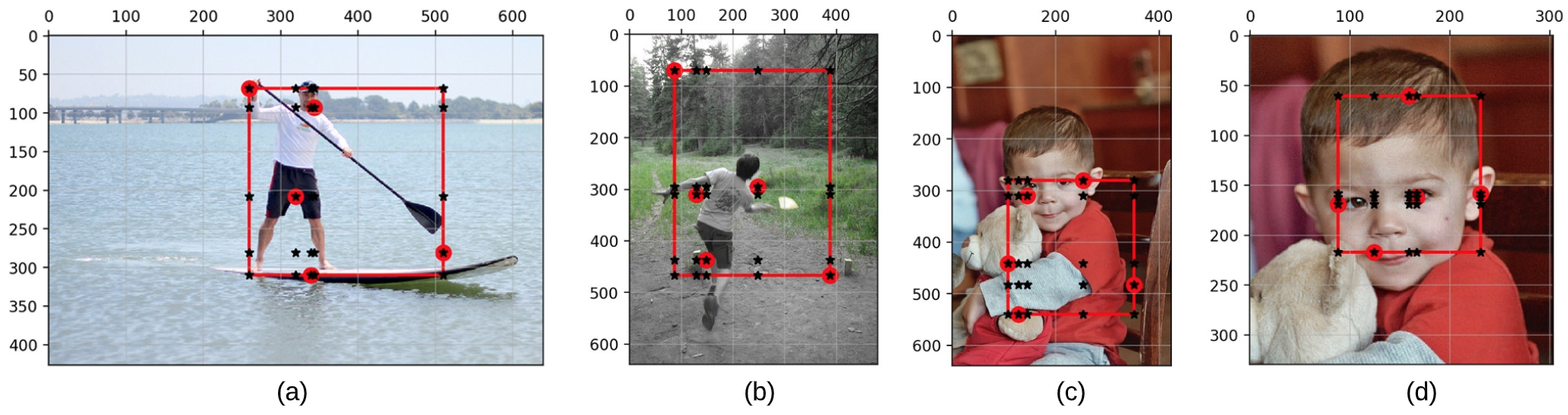}
\end{center}
  \caption{\textit{flexgrid2vec} generates Flexible Grid-Graphs for different image positions.}
\label{people}
\end{figure}

Visual recognition methods aim to learn discriminative image features that are useful in different downstream tasks, such as image classification. Existing methods, e.g. CNN-based, suffer from producing poor representation vectors due to challenges such as noise backgrounds and unrepresentative descriptors. For examples, Fig. \ref{people} shows different images where the background covers large spans in (a and b) or shares colours with the foreground object in (c and d). Moreover, in Fig. \ref{GG} (a), the background has no useful information about the image. In this paper, we define the best visual representation method to learn only the important patches and their spatial relationships in an image. We propose \textit{flexgrid2vec}, a novel GCN architecture to identify the important \textit{local} patches and connect them through a graph to capture the \textit{global} information.

\section{Architecture: \textit{flexgrid2vec}}\label{proposal}
We propose, \emph{flexgrid2vec}, an algorithmic framework for visual representation learning. It aims to learn CNN features through a novel GCN architecture. We consider an image $I$ as a grid-graph $G = (V, E)$ that consists of a set of nodes (image patches) and edges. The node coordinates and edges distances are flexible, i.e. dynamically changing from one image to another. \textit{flexgrid2vec} converts this grid-graph into a discriminative feature vector $f(I)$ as follows:
\begin{equation}\label{vector}
  f(I) = \sum_{v \in V} \sum_{e \in E(v)} \varphi(v, e)
\end{equation}
where $E(v)$ denotes the edges of the current node and $\varphi(v,e)$ is a GCN network that aggregates the CNN features from the neighbour nodes based on the grid connecting edges $E$. Table \ref{notations} lists the used notations. The $\varphi(v,e)$ is implemented to compute $H \in R^{n\times b}$, the node embeddings considering different edge attributes as in Eq. \ref{g_embd}.
\textit{flexgrid2vec} includes three main steps:
\begin{enumerate}
  \item Grid-graph construction.
    \item Feature extraction through multiple CNNs. 
    \item Embeddings learning via GCN.
\end{enumerate}

\subsection{Grid-graph Construction}
\begin{figure}[!ht] 
\begin{center}
  \includegraphics[angle=0, origin=c, width=0.8\linewidth]{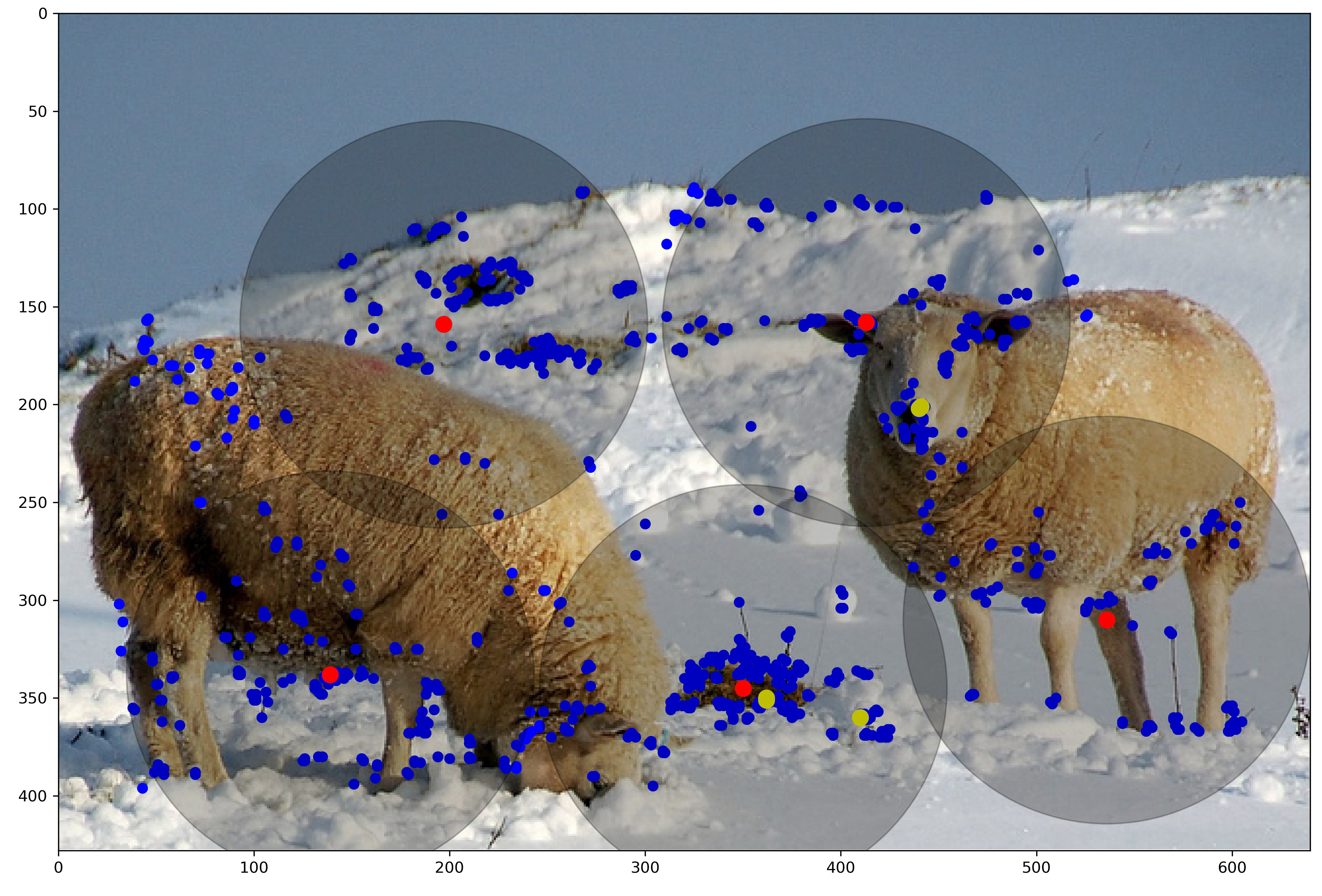}
\end{center}
  \caption{The proposed key-point detection method has two phases, 1) ORB detects a large set of high-quality points (blue) and 2) K-means clusters them into a few clusters (red). These final points (red) are significantly representing the different image regions. This can be compared with the ORB outcome of few points (yellow).}
\label{orb}
\end{figure}

\textit{flexgrid2vec} represents an input image in the form of a flexible grid-graph. We select candidate key-points (pixels) that will be used to initialise the graph node coordinates. Fig. \ref{GG} (a and b) visualise the detected key-points using the Oriented FAST and Rotated BRIEF ORB algorithm \cite{rublee2011orb}. ORB uses a patch intensity centroid to compute corner orientation, according to Eq. \ref{orb1} and \ref{orb2}, where $I(x,y)$ is the image intensity at the coordinate $(x,y)$, $p\ and\ q$ are the descriptive image moments and $C$ is the centroid. 
\begin{equation}\label{orb1}
  m_{pq} = \sum_{x,y} x^py^qI(x,y)
\end{equation}
\begin{equation}\label{orb2}
  C = \left (\frac{m_{10}}{m_{00}},\frac{m_{01}}{m_{00}}\right )
\end{equation}
where $m_{00}$ denotes the area or volume of the image pixels, $m_{10}$ is the sum over $x$ and $m_{01}$ refers to the sum over $y$. 
We use ORB to select a large set of pixels that covers the whole image. In some cases, ORB is not able to find enough key-points. Thus, a random set of pixels is used. Then, we utilise the K-means clustering approach to group this large set of pixels into a small set, as in Fig. \ref{orb}. The selected pixels are used as the graph nodes. The coordinates of these nodes contribute to the graph \textit{flexibility} in the same class through various images. For example, Fig. \ref{people} (a, b, c and d) represent the graphs for different people while having unique spatial distributions of the nodes. They also contribute to the scene identities. For example, although human faces in Fig. \ref{GG} (a) and \ref{people} (d) have quite different graph representations, they are still similar in comparison to the full-body graphs in Fig. \ref{people} (a, b and c). In the next stage, we generate image patches for the graph nodes (selected pixels). Rublee et al. (ICCV 2011) report experiment results that support our selection, as follows:
\begin{itemize}
  \item Detection time per frame (ms): ORB (15.3), SURF (217.3) and SIFT (5228.7).
    \item ORB had the best detection performance on synthetic data with more than 70\% inliers, followed by SIFT, which dropped by 10\%.
    \item On a real outdoor dataset: ORB (45.8\%), SURF (28.6\%) and SIFT (30.2\%).
\end{itemize}

\subsection{Feature extraction through multiple CNNs}
This work aims to reduce the learning of visual space to the most representative image patches. \textit{flexgrid2vec} extracts the features for only a pre-defined small number of patches. These patches are centred around the above-mentioned key-point detection method (i.e. combining ORB and clustering). We slice a square patch of pixels around the centre of each node, as shown in Fig. \ref{GG} (f). This process tends to eliminate the insignificant or unrepresentative image parts, as can be seen in Fig. \ref{GG} (c). The generated patches represent all different essential regions in the original image in Fig. \ref{GG} (a). The overall size of the generated patches is tiny compared to the size of the original image. 

\textit{flexgrid2vec} then extracts convolution features for each patch. We utilise a CNN pre-trained on the ImageNet. This network can be replaced by any traditional image feature extraction method, such as histograms of gradients or colours. The extracted convolutional features are embedded in their node in a singularity model. These node features represent the graph nodes' attributes used to compute the node embedding $H$.

\begin{figure*}[!ht] 
\begin{center}
  \includegraphics[angle=0, origin=c, width=0.99\linewidth]{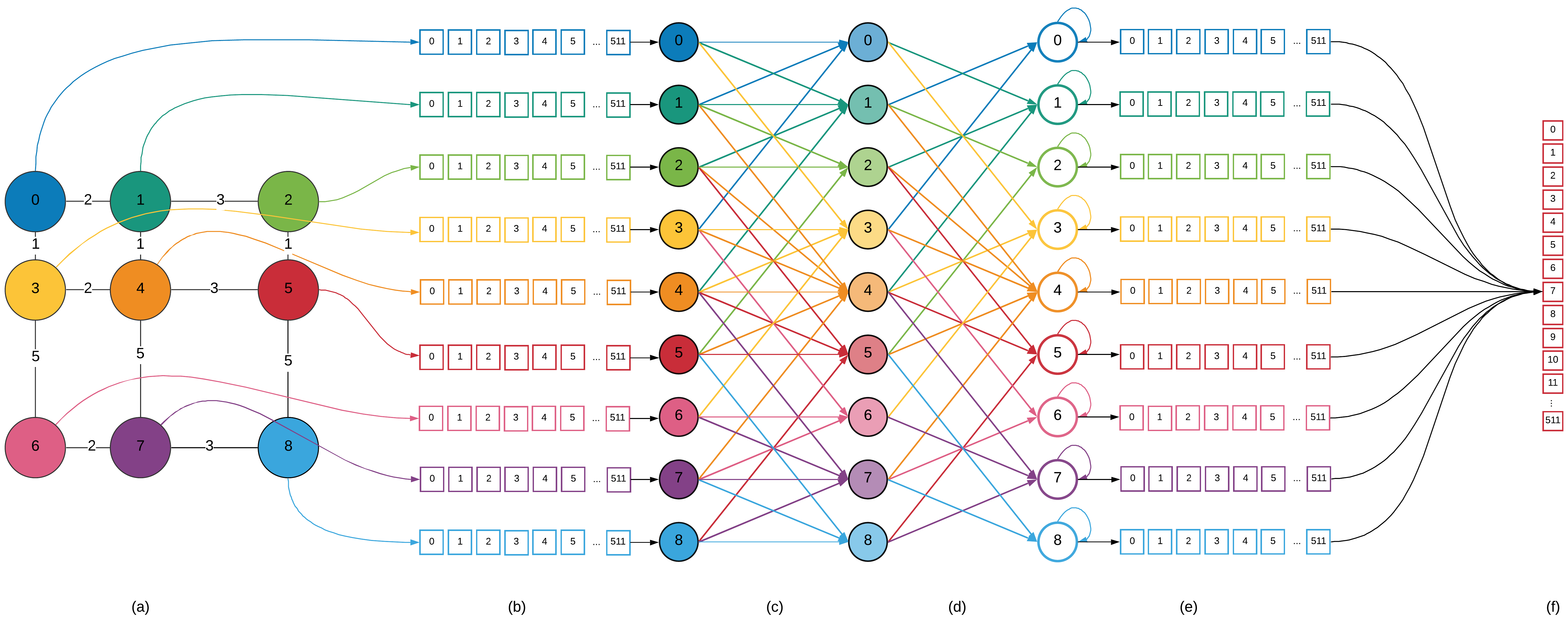}
\end{center}
  \caption{\textit{flexgrid2vec} node aggregation. (a) shows an example of a nine-node image. (b) CNN features as input to the embedding network. (c) aggregating the features of each node’s neighbours. (d and e) calculate the output vector of each node. (f) the sum of all graph vectors into the final representation vector.}
\label{node_aggregation}
\end{figure*}

\subsection{Embeddings Learning via GCN}
This section explains the general GCN architecture used in the literature and the proposed \textit{flexgrid2vec} GCN architecture. 
\subsubsection{GCN.}
The learning process of GCN depends on two functions, including node embedding and output learning. GCN aims to learn the graph $G$ embeddings as $H \in R^{n\times b}$, which comprises the neighbourhood information of the graph nodes $V$. The state $h_{v} \in {R}^{b}$ represents the node $v$ embedding in a \emph{b}-dimension vector. This process can be defined as a parametric local transition function. This function is shared among all nodes and updates the node embeddings based on the input from the neighbourhood. Specifically, this function computes the node embedding using the node's features, edges and neighbour information. The second main function is to learn the output label of each node. The embedding and output functions can be defined as follows:\begin{equation}\label{embd_network}
  h_{v} = \sum_{u \in N(v)} f(x_{v}, x^e_{(v,u)}, h_u, x_u)
\end{equation}
where $h_{v}$ is the embedding state of the node $v$, $x_{v}$ are the node features, $x^e_{v,u}$ the feature vector of the edge between the node $v$ and its neighbour $u$, and $h_u$ and $x_u$ are the embeddings and features of neighbourhood nodes. 
The aggregation of these embeddings can be computed as follows:
\begin{equation}\label{g_embd}
  h_{v} =  X_{v} \frac{\eta_{n}^{T}}{\varepsilon_{v}} + \sum_{u \in N(v)} X_{u} \frac{\eta_{u}^{T}}{\varepsilon_{u}} + \sum_{z \in N(u)} X_{z} 
    \left ( \frac{\eta_{z}^{T}}{\varepsilon_{z}} \right )^\frac{1}{2}
\end{equation}
where $X_{n}$ denotes the CNN features of the node patch, $\eta_{u}^{T}$ is the transpose of the eigenvector of the node, $X_{u}$ represents CNN-based features of the node neighbour $N(v)$, $\varepsilon$ is the eigenvalue of the current node, $N(u)$ represents the neighbours of the current node's neighbours and $X_z$ is the features of the neighbour $z$. 
One possible solution to compute the eigenvectors is to use a Fourier basis. Fourier basis eigenvectors and their corresponding eigenvalues represent the direction and variance of the graph Laplacian. The decomposed eigenvectors are a matrix $U \in R^{n \times n}$ that contains $\eta_{n}$, where $n$ dimension is the same as the node counts. These eigenvectors have a natural signal-frequency interpretation for the graph. The spectral Fourier basis decomposition produces the matrix \textit{U} that diagonalises the Laplacian as follows:
\begin{equation}
  L = diag (A + I)^{-1} = U \Lambda U^{T}
\end{equation}
where $\Lambda$ is a diagonal matrix of non-negative real eigenvalues. However, the calculation of the Fourier basis eigenvectors requires high-cost computations. Therefore, we propose to utilise the eigenvector centrality scores.

Then, the output function is computed as follows:
\begin{equation}\label{out_network}
  o_{v} = g(h_{v}, x_{v})
\end{equation}
where $g$ is a local function that describes the node output label. By stacking the versions of $f$ and $g$ for all nodes, we get $F$ and $G$, which are the global transition and output functions. GCN uses a gradient-descent algorithm for the learning process with $loss = \sum_{i \in n}(\large y_{i} - \large \hat y_{i})$, where $n$ is the number of labelled nodes and $\large y$ and $\large \hat y$ represent the target and output labels, respectively. Finally, the learnt node embeddings are concatenated in one vector to represent the given image. However, this generic GCN architecture is not applicable in our use-case. To represent an image as a vector, we do not have different node labels. Therefore, we model the GCN on our proposed \textit{flexgrid2vec}, as explained below.

\subsubsection{The proposed GCN.}
\textit{flexgrid2vec} defines the grid-graph as $G = (V, E)$ of nodes \textit{V} and edges \textit{E}. The goal of the proposed GCN is to learn the embedding state $h_{v} \in {R}^{b}$. \textit{flexgrid2vec} uses the node attributes, i.e. the CNN-based extracted features, to learn the node embeddings $H \in R^{n\times b}$. \textit{flexgrid2vec} constructs the grid-graph as a flexible skeleton where the graph edges are attributed by the distance between the nodes as $E_{v,u} = distance (v, u)$. Therefore, the edges mostly have different lengths. 

To this end, we have defined the grid-graphs, its nodes and node features, and its edges and edge attributes. Next, we define each node's neighbourhood and compute the node embeddings $H \in R^{n\times b}$. \textit{flexgrid2vec} computes the node embedding based on the node CNN features, as follows:
\begin{equation}\label{embding}
  h_{v} =  X_{v} + \sum_{u \in N(v)} X_{u} \frac{\gamma}{\theta} + \sum_{z \in N(u)} X_{z} \frac{\gamma}{\theta}
\end{equation}
where $X_{v}$ and $X_{u}$ denote the CNN features of the current node patch and the node neighbours $N(v)$, $N(u)$ represents the neighbours of neighbours, $X_z$ are the features of the neighbour and $\frac{\gamma}{\theta}$ is a normalisation factor. This normalisation process aims to scale the embedding features into a specific range. The $\gamma$ and $\theta$ variables can be pre-defined or computed based on the spectral graph theory. 

We investigate the impact of using the spectral graph components on the embedding vectors. The embedding computing methods in Eq. \ref{embding} will be updated as follows:
\begin{equation}\label{embdingc}
  h_{v} =  X_{v} c_v + \sum_{u \in N(v)} \frac{X_{u} c_u}{\theta} + \sum_{z \in N(u)} \frac{X_{z} c_z}{\theta}
\end{equation}
where $c_v$ and $c_u$ are the eigenvector centrality scores of the node and its neighbours.
Eigenvector centrality calculates the centrality score $C$ for each node based on its neighbours. Eigenvector centrality is an extension of the simple degree centrality. Degree centrality is an awarding mechanism giving one centrality point for every neighbour in the graph. Eigenvector centrality considers the importance level of each neighbour. The node importance increases or decreases based on the importance of its neighbours. Eigenvector centrality assigns each node a score proportional to the sum of its neighbours' scores as follows:
\begin{equation}
  c_v = \varepsilon_l^{-1} \sum A_{v,u}c_u
\end{equation}
where $c_v$ represents the centrality of the node $v$, $\varepsilon_l$ is the largest eigenvalue, $A_{v,u}$ is an element in the adjacency matrix and $c_u$ is the neighbour node. In this scenario, for graph convolution, Eq. \ref{embding} concatenates \textit{+} the aggregated sum of the node neighbours and their neighbours. This operation limits the convolution as a two-step aggregation process. In this paper, we will test different versions of Eq. \ref{embding} and \ref{embdingc}. 

To this end, we introduced the main components of \textit{flexgrid2vec}, including the construction of the Flexible Grid-Graphs and learning the node embeddings. In the following sections, we discuss the experiment results for each component. 

\section{Experiment Settings}\label{experimental}

\subsection{Implementation Details}
We used multiple Python packages to implement the different components of \textit{flexgrid2vec}, including, 1) Grid-graph construction, 2) Feature extraction through multiple CNNs and 3) Embeddings learning via GCN. We conducted all experiments in a HPE NVIDIA Tesla $V100$ GPU server with $512GB$ memory.

The following are the model parameter settings:
\begin{itemize}
  \item \textbf{Key-point detection.} The key-point detection method is parameterised by two variables, including the ORB key-point numbers and the output clusters (key-pixels). We specified the feature number to be $250$, as shown in blue in Fig. \ref{orb}. Then, we used K-means clustering to select the most representative $5$ clusters, i.e. points, as visualised in red in Fig. \ref{GG} (c).
    \item \textbf{Grid-graph construction.} We constructed a grid-graph based on the coordinates of the detected key-pixels.
    \item \textbf{Patch Generation}. We generated $5\times 5$ patches of size $32 \times 32$ centred around the detected key-pixels. The generated patches were a total of $25$ patches ($25,600$ pixels) instead of the original image-size of $438 \times 640$ ($280,320$ pixels). The pixel rate was reduced by around 91\% of the original image.
    \item \textbf{CNN Feature Extraction}. We employed the TensorFlow pre-trained VGG model to extract the $X_v$ convolutional features for each $v$ node.
    .
    \item \textbf{GCN embedding setting}. The most optimal setting for the GCN embedding is aggregating the node CNN features on a one-step neighbourhood, denoted as \textit{flexgrid2vec-Agg1R}. We have investigated multiple node-embedding methods as explained below. 
\end{itemize}

\subsection{Different Embedding Configurations}

We implemented \textit{flexgrid2vec} using various embedding aggregation methods. These methods are as follows:
\begin{enumerate}
  \item \textit{flexgrid2vec-Agg1R} computes the node feature embeddings by aggregating the direct connected neighbours. It then takes the sum of the $25$ node vectors to represent the image in a $1 \times 512$ feature vector. The node embedding method in Eq. \ref{embding} will be defined as follows: $h_{v} =  X_{v} + \sum_{u \in N(v)} \frac{X_{u}}{\theta}$.
    \item \textit{flexgrid2vec-Agg1} computes the node feature embeddings by aggregating the direct connected neighbours. The aggregated vectors are concatenated as one vector. For example, a $5 \times 5$ grid-graph that has $25$ nodes will have $25 \times 512$ features concatenated in the output vector. The node embedding method in Eq. \ref{embding} will be defined as follows: 
    $h_{v} =  X_{v} \frown \frac{X_{u}}{\theta} \forall {u \in N(v)} $.
    \item \textit{flexgrid2vec-Agg2R} computes the sum of the vectors of the current node's neighbours and their neighbours. It follows the same node embedding method in Eq. \ref{embding}.
    \item \textit{flexgrid2vec-Agg2} works like \textit{flexgrid2vec-Agg2R} but with concatenation instead of summation. The node embedding method in Eq. \ref{embding} will be defined as follows: $h_{v} =  X_{v} \frown \frac{X_{u}}{\theta} \forall{u \in N(v)}  \frown \sum_{z \in N(u)} \frac{X_{z}}{\theta}$.
    \item \textit{flexgrid2vec-EVC1R} normalises the output of \textit{flexgrid2vec-Agg1R} with the eigenvector centrality score. The node embedding method in Eq. \ref{embdingc} will be defined as follows: $h_{v} =  X_{v} c_v + \sum_{u \in N(v)} \frac{X_{u} c_u}{\theta}$.
    
    \item \textit{flexgrid2vec-EVC1} normalises the output of \textit{flexgrid2vec-Agg1} with the eigenvector centrality of the Flexible Grid-Graphs. The node embedding method in Eq. \ref{embdingc} will be defined as follows: 
    $h_{v} =  X_{v} c_v \frown \frac{X_{u} c_u}{\theta} \forall{u \in N(v)}$.
    \item \textit{flexgrid2vec-EVC2R} considers the two-step neighbour aggregation, normalisation with eigenvector centrality and vector size reduction via summation. It follows the same node embedding method as Eq. \ref{embdingc}.
    \item \textit{flexgrid2vec-EVC2} extends the method of \textit{flexgrid2vec-EVC1} to the neighbours of neighbours at only two steps. The node embedding method in Eq. \ref{embdingc} will be defined as follows: 
    $h_{v} =  X_{v} c_v \frown \frac{X_{u} c_u}{\theta} \forall{u \in N(v)} \frown \sum_{z \in N(u)} \frac{X_{z} c_z}{\theta}$.
\end{enumerate}
Fig. \ref{node_aggregation} shows the main steps of the node aggregation process.

\subsection{Datasets}\label{datasets}
Two datasets are considered to conduct experiments with \textit{flexgrid2vec} on both visual representation learning and image classification tasks. Specifically, we used: 
\begin{enumerate}
  \item CIFAR-10 \cite{krizhevsky2009learning} consists of $60,000$ images. The images are $32 \times 32$ in size and divided into $50,000$ for training and $10,000$ for testing. CIFAR-10 has $10$ categories, such as aeroplane, automobile, bird, cat, deer, dog and frog. The categories are mutually exclusive with no overlapping. 
    \item CIFAR-100 \cite{krizhevsky2009learning} is similar to the CIFAR-10 but has $100$ classes containing 600 images each.
    \item STL-10 \cite{coates2011analysis} was prepared for image recognition model evaluation. STL-10 has fewer labelled training examples and has $5,000$ images for training and $8,000$ for testing, over $10$ classes, as well as $100,000$ unlabelled images for unsupervised learning. However, in this paper, we did not use any of these unlabelled images. The STL-10 $5,000$ and $8,000$ images are $96 \times 96$ and are acquired from the ImageNet.
    \item ASIRRA (Animal Species Image Recognition for Restricting Access), published by Microsoft for binary classification, is a balanced dataset having, for each class, $12,500$ and $1,000$ images for training and testing, respectively.
    \item COCO (Microsoft Common Objects in Context) dataset \cite{lin2014microsoft} for multi-class image classification has $123,287$ images annotated for $80$ object categories and $12$ super categories. One image may be annotated to one or more object classes. We filter the dataset to images that are only annotated to one class. Therefore, the final utilised dataset is around $29,000$ images for multi-class classification on 12 super categories. This utilised sub-set produces challenging issues of imbalanced and heterogeneous images. This heterogeneity stemmed from the utilisation of the super high-level categories: vehicles, indoor and accessories. The heterogeneity issue can be seen in multiple ways: different backgrounds as in the vehicle category, e.g. sea and grass; miscellaneous items as in the indoor class, e.g. vase and teeth brush; and various types as in accessories, e.g. bag and umbrella.
\end{enumerate}

\subsection{Baseline and State-of-the-art Methods}
We compare the image classification results of  \textit{flexgrid2vec} with state-of-the-art CNN-based methods that are pre-trained on ImageNet, such as VGG \cite{simonyan2014very}, ResNets \cite{he2016deep,huang2016deep,he2016identity}, DenseNet \cite{huang2017densely}, MobileNet \cite{Sandler_2018_CVPR,howard2017mobilenets}, NASNet \cite{zoph2018learning}, Xception \cite{chollet2017xception} and InceptionV3 \cite{szegedy2016rethinking}. 
We also compare the embedding method of the proposed \textit{flexgrid2vec} with the existing embedding models:  node2vec \cite{grover2016node2vec} and GCN-based aggregated nodes \cite{kipf2017semi}. In addition, we compare flexgrid2vec's accuracy with recent algorithms, such as MixMatch \cite{NEURIPS2019_MixMatch} (NIPS 2019), DIANet \cite{huang2020dianet} (AAAI2020), reSGHMC \cite{deng2020non} (ICML 2020), Prodpoly \cite{chrysos2021deep} (TPAMI 2021), ACNet \cite{wang2019adaptively} (CVPR 2019), Quadratic-Embedding \cite{Cordonnier2020On} (ICLR 2020) and CLS-GAN \cite{qi2020loss}, (IJCV 2020). 

\section{Evaluation}\label{results}
For evaluation purposes, we utilised \textit{flexgrid2vec} to extract the feature vectors of two image datasets. Then, we developed a simple fully connected neural network to classify the images. In the following sub-sections, we discuss the performance of \textit{flexgrid2vec} to show its efficiency and accuracy. We test \textit{flexgrid2vec} as a visual representation algorithm. We visualise the produced vectors to highlight the learnt feature space. We also tested \textit{flexgrid2vec} on binary and multi-class image classification tasks. Finally, we evaluate \textit{flexgrid2vec} under different configurations and with multiple node-embedding methods. The architecture with the \textit{AGG1R} embedding method showed the best results. Therefore, we refer to that version by \textit{flexgrid2vec} in the rest of this paper. 

\begin{table}[]
\begin{center}
\caption{Classification accuracy (top 1) results on CIFAR-10.}\label{CIFAR-10}
\small
\centering
\begin{tabular}{|p{3.5cm}|l|l|}
\hline
Model & Test Accuracy & Venue (Year)\\ \hline
  DeepInfoMax (G) \cite{hjelm2018learning} & 52.84\% & ICLR 2019\\ \hline
  Adversarial AutoEncoder \cite{makhzani2015adversarial} & 57.19\% & ICLR 2016 \\ \hline
  Variational AutoEncoder (VAE) \cite{kingma2013auto} & 60.54\% & \\ \hline
ANODE \cite{NIPS2019_8577} & 60.6\% & NeurIPS 2019 \\ \hline %
  BiGAN \cite{makhzani2015adversarial,dumoulin2016adversarially} & 62.74\% & \\ \hline
DeepInfoMax (infoNCE) \cite{hjelm2018learning} & 75.57\% & ICLR 2019\\ \hline %
DenseNet \cite{huang2017densely} & 77.79\%  & CVPR 2017\\ \hline %
  DenseNet (Spinal FC) & 81.13\% & \\ \hline
DCGAN \cite{yu2017unsupervised} & 82.8\% &  \\ \hline
  Roto-Scat + SVM \cite{oyallon2015deep} & 82.30\%  & \\ \hline
  ExemplarCNN \cite{dosovitskiy2014discriminative} & 84.3\%  & \\ \hline
Baikal \cite{gonzalez2020improved} & 84.53\%  & CEC 2020\\ \hline %
CapsNet \cite{sabour2017dynamic} & 89.4\%  & NeurIPS 2017\\ \hline %
MP \cite{hendrycks2016baseline} & 89.07\% & \\ \hline

MIM \cite{liao2016importance} & 91.5\%  & WACV 2016\\ \hline %
CLS-GAN \cite{qi2020loss} & 91.7\%  & IJCV'2020\\ \hline %
DSN \cite{lee2015deeply} & 91.8\%  & PMLR'15\\ \hline %
BinaryConnect \cite{courbariaux2015binaryconnect} & 91.7\%  & NeurIPS 2015\\ \hline %
Mish \cite{misra2020mish} & 92.20\%  & BMVC 2020\\ \hline %

SA quadratic embedding \cite{Cordonnier2020On} & 93.8\%  & ICLR 2020\\ \hline %
Standard ACNet \cite{wang2019adaptively} & 94\%  & CVPR 2019\\ \hline %
Deep Complex \cite{trabelsi2018deep} & 94.4\%  & ICLR 2018 \\\hline %

Prodpoly \cite{chrysos2021deep} & 94.9\% & TPAMI 2021 \\ \hline %
MixMatch \cite{NEURIPS2019_MixMatch} & 95.05\% & NEURIPS 2019 \\ \hline %
SRM-ResNet-56 \cite{lee2019srm} & 95.05\% & ICCV 2019 \\ \hline %
Evolution ensemble \cite{real2017large} & 95.6\%  & ICML 2017\\ \hline %
PreActResNet18+AMP \cite{zheng_2021_CVPR} & 96.03\%  & CVPR 2021 \\ \hline %
Wide Residual Networks \cite{zagoruyko2016wide} & 96.11\% &  \\ \hline %
ResNet20 with reSGHMC \cite{deng2020non} & 94.62\%  & ICML 2020\\ \hline %
ResNet56 with reSGHMC \cite{deng2020non} & 96.12\%  & ICML 2020\\ \hline %

\textbf{flexgrid2vec} & \textbf{96.23\%} &  \\ \hline
\end{tabular}
\end{center}
\end{table}

\subsection{Benchmarking Results}
We compare the benchmarking results of \textit{flexgrid2vec} with the state-of-the-art methods. Although \textit{flexgrid2vec} produces significantly reduced feature vectors, it has high classification accuracy. Tables \ref{CIFAR-10}, \ref{CIFAR-100}, \ref{STL-10}, \ref{binary} and \ref{mc} list the learning test accuracy. 

\textit{flexgrid2vec} achieves $96.23\%$, outperforming baseline and state-of-the-art image classification methods, as listed in Table \ref{CIFAR-10}. As \textit{flexgrid2vec} is a batch-based visual representation, it is directly compared to DeepInfoMax \cite{hjelm2018learning}. DeepInfoMax has been implemented based on different configurations and loss functions. It has $52.84\%$, $70.60\%$, $73.62\%$ and $75.57\%$ accuracy. Also, \textit{flexgrid2vec} outperforms recent advances in the state-of-the-art AutoEncoder (AE), such as Adversarial AE, Variational AE and $\beta$-VAE, which have achieved $57.19\%$, $57.89\%$ and $60.54\%$, respectively. Table \ref{CIFAR-100} shows the experiment results on the CIFAR-100 dataset. \textit{flexgrid2vec} achieves $81.47\%$, outperforming recent methods, such as ResNet-1001 \cite{he2016identity}, MixMatch \cite{NEURIPS2019_MixMatch}, Mish \cite{misra2020mish} and DIANet \cite{huang2020dianet}. On the other hand, the benchmark results on the STL-10 dataset show that the proposed \textit{flexgrid2vec} outperforms different methodologies, such as GAN \cite{makhzani2015adversarial}, Autoencoders \cite{makhzani2015adversarial,kingma2013auto,higgins2016beta,alemi2016deep,dumoulin2016adversarially} and patch-based DeepInfoMax \cite{hjelm2018learning}.

In binary classification on the ASIRRA dataset, \textit{flexgrid2vec} achieves $98.8\%$, outperforming all the state-of-the-art methods. In multi-class, on the COCO dataset, \textit{flexgrid2vec} achieves $89.69\%$, having the first rank with the VGG19. The results of both binary and multi-class image classifications show the consistency of \textit{flexgrid2vec}'s learning behaviour. \textit{grdi2vec} learns smoothly in both tasks. Although some state-of-the-art algorithms achieve high accuracy, they show overfitting and fluctuation between the train and test results. For example, NASNetMobile has $98$\% in training and only $47$\% in testing. It appears they suffer from the challenging issues of the utilised dataset as discussed in the datasets section. This result confirms the superiority of \textit{flexgrid2vec} as an efficient visual representation learning approach.

\begin{table}[!h]
\begin{center}
\caption{Classification accuracy (top 1) results on CIFAR-100.}\label{CIFAR-100}
\small
\begin{tabular}{|p{3.5cm}|l|l|}
\hline
Model & Test Accuracy & Venue (year)\\ \hline
DSN \cite{lee2015deeply} & 65.4\% & PMLR 2015\\ \hline 
Unsharp Masking \cite{carranza2019unsharp} & 60.36\% & ICANN 2019\\ \hline 
ResNet-50  \cite{he2016identity} & 67.06\% & PMLR 2015 \\ \hline 
GFE \cite{Hertel2015Deep} & 67.7\% & IJCNN 2015 \\ \hline 
MIM \cite{liao2016importance} & 70.8\% & WACV 2016 \\ \hline
MixMatch \cite{NEURIPS2019_MixMatch} & 74.10\% & NIPS 2019 \\ \hline 
Mish \cite{misra2020mish} & 74.41\% & BMVC 2020 \\ \hline 
Stochastic Depth \cite{huang2016deep} & 75.42\% & ECCV 2016 \\ \hline 
Exponential Linear Units \cite{clevert2015fast} & 75.7\% & \\ \hline 
DIANet \cite{huang2020dianet} & 76.98\% & AAAI2020 \\ \hline 
Evolution \cite{real2017large} & 77\% & ICLR 2017 \\ \hline 
ResNet-1001 \cite{he2016identity} & 77.3\% & ECCV 2016 \\ \hline 
\textbf{flexgrid2vec} & \textbf{83.05\%} & \\ \hline
\end{tabular}
\end{center}
\end{table}

\begin{table}[!h]
\caption{Classification accuracy (top 1) results on STL-10.}\label{STL-10}
\small
\centering
\begin{tabular}{|p{3.5cm}|l|l|}
\hline
Model & Test Acc.  & Venue (Year) \\ \hline
Variational AutoEncoder (VAE) \cite{kingma2013auto} & 56.72\% & ICLR 2014\\ \hline
AE & 55.57 & \\ \hline
$\beta$-VAE \cite{higgins2016beta} & 55.14\% & ICLR 2017\\ \hline
Adversarial Autoencoders (AAE) \cite{makhzani2015adversarial} & 54.47\% & ICLR 2016 \\ \hline
Noise As Target (NAT) \cite{bojanowski2017unsupervised} & 61.43\% & \\ \hline
DeepInfoMax (DV) \cite{hjelm2018learning} & 63.81\% &  ICLR 2019\\ \hline

Accuracy Monitoring \cite{shao2020increasing} & 63.43\% & IJCAI-PRICAI 2020 \\ \hline
Hierarchical Matching Pursuit (HMP)\cite{oyallon2017scaling} & 64.5\%  & ICCV 2017\\ \hline
BiGAN \cite{dumoulin2016adversarially} & 67.18\% & \\ \hline
DFF Committees \cite{Miclut_2014} & 68\% & Pattern Recognition 2014 \\ \hline
C-SVDDNet \cite{Wang2016C-SVDDNet} & 68.20\% & Pattern Recognition 2016\\ \hline
DeepInfoMax (JSD) \cite{hjelm2018learning} & 70.85\% &  ICLR 2018\\ \hline
ResNet18 \cite{luo2020extended} & 72.66\%  & \\ \hline
Second-order Hyperbolic CNN \cite{ruthotto2019deep} & 74.3\%  & JMIV 2019 \\ \hline
Exemplar CNN \cite{oyallon2017scaling} & 75.7\% & ICCV 2017 \\ \hline %
Parabolic \cite{ruthotto2019deep} & 77\% & JMIV 2019 \\ \hline %
PSLR-Linear \cite{Ye_2020_CVPR} & 78.8\% & CVPR 2020 \\ \hline %
Greedy InfoMax \cite{LoweOV19NeurIPS} & 81.9\% & NeurIPS 2019 \\ \hline %
ResNet baseline \cite{robert2018hybridnet} & 82\% & ECCV 2018 \\ \hline %
PSLR-knn \cite{ye2020probabilistic} & 83.2\% & CVPR 2020 \\ \hline %
HybridNet \cite{robert2018hybridnet} & 84.10\% & ECCV 2018 \\ \hline %
MidPoint \cite{chang2018reversible} & 84.6\% & AAAI 2018 \\ \hline %
IIC \cite{ji2019invariant} & 88.8\% & ICCV 2019 \\ \hline %
SESN \cite{Sosnovik2020Scale} & 91.49\% & ICLR 2020 \\ \hline %
NSGANetV2 \cite{lu2020nsganetv2} & 92.0\% & ECCV 2020 \\ \hline %
FixMatch \cite{NEURIPS20_FixMatch} & 92.34\% & NEURIPS 2020 \\ \hline %
AMDIM \cite{chen2020generative} & 93.80\% & ICML 2020 \\ \hline %
ReMixMatch \cite{NEURIPS2019_MixMatch} & 93.82\% & ICLR 2019 \\ \hline %
AMDIM-L \cite{chen2020generative} & 94.2\% & ICML 2020 \\ \hline %
MixMatch \cite{NEURIPS2019_MixMatch} & 95.05\% & NEURIPS 2019 \\ \hline %

\textbf{flexgrid2vec (Ours)} &\textbf{94.50} & \\ \hline
\end{tabular}
\end{table}

\begin{table}[!h]
\begin{center}
\caption{Binary classification benchmarking results on ASIRRA dataset.}\label{binary}
\small
\begin{tabular}{|p{3cm}|l|l|}
\hline
Model & Test Accuracy & Test Loss \\ \hline
VGG16  \cite{simonyan2014very} & 82.2\% & 0.274 \\ \hline
VGG19 \cite{simonyan2014very} & 88.2\% & 0.137 \\ \hline
ResNet50 \cite{he2016deep} & 50\% & 0.502 \\ \hline
InceptionV3 \cite{szegedy2016rethinking} & 97.2\% & 0.032 \\ \hline
DenseNet121  \cite{huang2017densely} & 94.3\% & 0.073 \\ \hline
MobileNet \cite{howard2017mobilenets} & 98.4\% & 0.023 \\ \hline
NASNetMobile \cite{zoph2018learning} & 97.9\% & 0.022 \\ \hline
\textbf{flexgrid2vec (ours)} &
\textbf{98.8\%} & 
\textbf{0.049} \\ \hline
\end{tabular}
\end{center}
\end{table}

\begin{table}[!h]
\begin{center}
\caption{Multi-class classification benchmarking results on COCO dataset.}\label{mc}
\small
\begin{tabular}{|p{3cm}|l|l|}
\hline
Model & Test Accuracy & Test Loss \\ \hline
VGG16 \cite{simonyan2014very} & 79.0\% & 0.738 \\ \hline
VGG19 \cite{simonyan2014very} & 82.9\% & 0.620 \\ \hline
ResNet50 \cite{he2016deep} & 61.26\% & 0.240 \\ \hline
InceptionV3 \cite{szegedy2016rethinking} & 60.5\% & 1.923 \\ \hline
DenseNet121 \cite{huang2017densely}  & 75.05\% & 0.597 \\ \hline
MobileNet \cite{howard2017mobilenets} & 70.7\% & 1.196 \\ \hline
NASNetMobile (depth multiplier=1.0) \cite{zoph2018learning} & 5928\% & 0.674 \\ \hline
NASNetMobile (depth multiplier=0.5) \cite{zoph2018learning} & 62.60\% & 0.574 \\ \hline
\textbf{flexgrid2vec (ours)} & 
\textbf{89.69\%} & 
\textbf{0.11} \\ \hline
\end{tabular}
\end{center}
\end{table}

\begin{table*}[!h]
\caption{Classification accuracy (top 1) results on different parameter settings.}\label{param}
\centering
\begin{tabular}{|l|l|l|l|l|l|}
\hline
Dataset & Image Size & Base Model & Dimension & Patch Size & Test Acc. \\ \hline
\multirow{3}{*}{$STL-10$} 
& 96 & ResNet50 & 4096 & 20 & 92.95\% \\ \cline{2-6} 
& 96 & EfficientNet & 18448 & 20 & 88.15\% \\ \cline{2-6} 
& 224 & ResNet50 & 4096 & 8 & 94.50\% \\ \cline{2-6} 
\hline
\multirow{4}{*}{$CIFAR-10$} 
& 32 & ResNet50 & 1016 & 32 & 87.90\% \\ \cline{2-6} 
& 96 & ResNet50 & 2048 & 32 & 96.23\% \\ \cline{2-6} 
& 96 & EfficientNet & 18448 & 32 & 93.98\% \\ \cline{2-6} 
& 96 & ResNet50 & 2048 & 20 & 95.73\% \\ \hline
\multirow{3}{*}{$CIFAR-100$} 
& 32 & ResNet50 & 1016 & 32 & 63.88\% \\ \cline{2-6} 
& 96 & ResNet50 & 2048 & 32 & 81.38\% \\ \cline{2-6} 
& 224 & ResNet50 & 2048 & 32 & 83.05\% \\ \hline
$ASIRRA$ & 224 & None & None & 32 & 98.8\% \\ \hline
\multirow{2}{*}{$COCO$}
& 400 & None & Noe & 32 & 82.38\% \\ \cline{2-6} 
& 400 & ResNet50 & 2048 & 32 & 89.74\% \\ \hline
\end{tabular}
\end{table*}

The proposed \textit{flexgrid2vec} distinguishes the most important patches to learn accurate representations that are effective with a small fraction of the image patches. In order to validate the proposed methodology, we utilise two high-resolution image sets, namely ASIRRA and COCO, achieving $98.8\%$ and $82.38\%$, respectively. In order to compare \textit{flexgrid2vec} with state-of-the-art algorithms on low-resolution datasets, such as CIFAR-10, CIFAR-100 and STL-10, we combined the feature vectors of \textit{flexgrid2vec} with other baseline models, as shown in Table \ref{param}. 
The table lists the results of using \textit{flexgrid2vec} within different ablations on the five utilised datasets. Using \textit{ResNet-50} in combination with \textit{flexgrid2vec} achieves $94.50\%$ and $82.38$ to $89.74\%$ accuracy using the STL-10 and COCO datasets, respectively.
Table \ref{param} also shows the impact of changing the image size. \textit{flexgrid2vec} achieves better results on the STL-10 dataset when scaling the images up from $96$ to $224$. Increasing the patch size in the case of CIFAR-10 also improves the accuracy from $95.73\%$ to $96.23\%$. Moreover, increasing both the image and patch sizes increases the accuracy of \textit{flexgrid2vec} by around $2\%$. This trend can also be seen in COCO accuracy, which is increased by around $7\%$. However, implementing \textit{flexgrid2vec} on the binary classification data achieves high accuracy without the need to be combined with other deep models. 

\begin{figure}[!h]
\centering
  \includegraphics[angle=0, origin=c, width=0.99\linewidth]{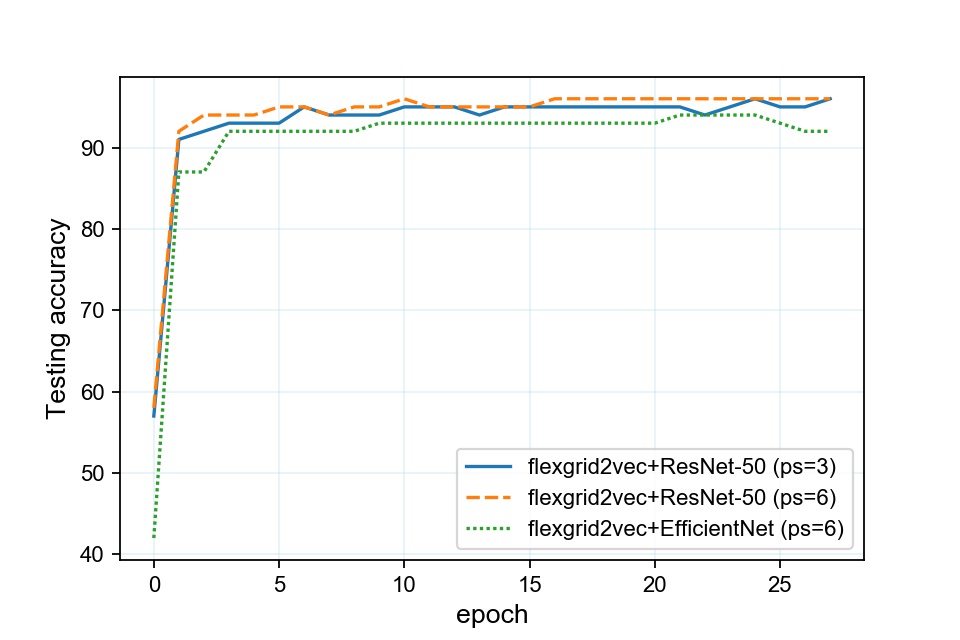}
  \caption{Testing accuracy of \textit{flexgrid2vec} on the CIFAR-10 dataset.}
\label{fg2v_CIFAR10}
\end{figure}

\begin{figure}[!h]
\centering
  \includegraphics[angle=0, origin=c, width=0.99\linewidth]{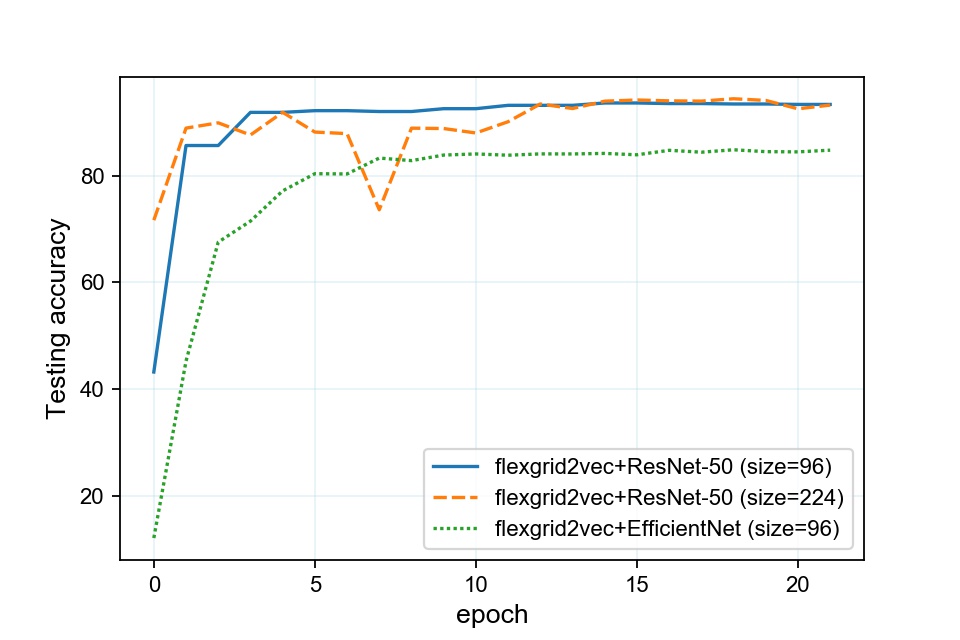}
  \caption{Testing accuracy of \textit{flexgrid2vec} on the STL-10 dataset.}
\label{fg2v_STL}
\end{figure}

Fig. \ref{fg2v_CIFAR10} and \ref{fg2v_STL} show the testing accuracy of \textit{flexgrid2vec} on the CIFAR-10 and STL-10 datasets. We experimented with CIFAR-10 using the proposed \textit{flexgrid2vec} in combination with ResNet-50 and EfficientNet with the same patch size of $32$ pixels. The combination with ResNet-50 outperformed the one with EfficientNet by $2.25\%$. We also reduced the patch size from $32$ to $20$ using the ResNet-50. The latter experiment shows that the $32$ setup has better accuracy, with $96.23\%$ in comparison to $95.73\%$ using the $20$ pixel patch setup. Fig. \ref{fg2v_STL} shows the impact of different image sizes using \textit{flexgrid2vec} and ResNet-50. Using the image size of $224$ pixels has the highest accuracy, though it fluctuates due to the large scaling from the original size of $32$ pixels. On the other hand, using an image size of $96$ pixels has less accuracy of $92.95\%$ but a steady learning curve. 


\subsection{Embedding Configuration}

\subsubsection{\textit{flexgrid2vec} Embedding Configuration Testing}
We conducted a large set of experiments to evaluate the performance of \textit{flexgrid2vec}. Table \ref{5kp_results} lists sample results of model training using different embedding methods. 
The embedding calculation methods have different learning behaviours. Specifically, the methods that concatenate the output vectors as one large vector tend to have inferior test accuracy, showing around $9$\% less than the training accuracy. However, reduced vector-based methods show better learning behaviour for both training and testing. This significant performance of the reduced \textit{flexgrid2vec} is caused by using the most discriminative $512$ features instead of using a large vector of $12,800$ features. 
The \textit{flexgrid2vec-Agg1R} that applies simple node aggregation and summation outperforms all other versions including the method with eigenvector centrality normalisation. This observation proves that the \textit{flexgrid2vec} requires lower computation to produce the best representation vector.
The average speed highlights how fast \textit{flexgrid2vec} is. The results show that \textit{flexgrid2vec} processes around $20$ images per second.

We evaluated the parameters of \textit{flexgrid2vec}, such as the patch size and node count. Table \ref{7kp_results} shows the performance results of building \textit{flexgrid2vec} grid-graphs on $7$ key-points ($7\times 7$ nodes). These results decreased by a range of $6$ to $15$ percent when using \textit{flexgrid2vec} on five key-points ($5\times 5$ nodes), see Table \ref{5kp_results}. Table \ref{64_results} shows the training results of the \textit{flexgrid2vec} reduced methods. These experiments were designed to utilise $64 \times 64$ patch sizes to calculate the convolutional feature of each node. The accuracy degraded significantly in all the methods. For example, the test accuracy of \textit{flexgrid2vec-Agg1R} decreased from $98.8$\% to $90.1$\%. This result supports the strength of the \textit{flexgrid2vec} proposed architecture that uses $32 \times 32$ patch sizes, see results in Table \ref{5kp_results}.

\begin{table}[!h]
\caption{Experiment results of using \textit{flexgrid2vec} embedding methods with patch size of $32 \times 32$.}\label{5kp_results}
\small
\center
\begin{tabular}{|l|l|l|l|}
\hline
Embedding Method & Test Acc. & Test loss & Time (s)  \\ \hline
\textbf{\textit{Agg1R}}  & \textbf{98.8\%} & \textbf{0.049} & \textbf{0.051} \\ \hline
\textit{Agg1}  & 88.7\% & 0.115 & 0.053 \\ \hline
\textit{Agg2}  & 90.8\% & 0.093 & 0.074 \\ \hline
\textit{Agg2R}  & 96.8\% & 0.102 & 0.070 \\ \hline
\textit{EVC1}  & 88.8\% & 0.303 & 0.053 \\ \hline
\textit{EVC1R}  & 97.9\% & 0.069 & 0.050 \\ \hline
\textit{EVC2}  & 90.2\% & 0.1 & 0.076 \\ \hline
\textit{EVC2R}  & 96.1\% & 0.122 & 0.076 \\ \hline
\end{tabular}
\end{table}

\begin{table}[!h]
\caption{Performance results of \textit{flexgrid2vec} with grid-graphs of $7 \times 7$ (49) nodes.}\label{7kp_results}
\small
\center
\begin{tabular}{|l|l|l|l|}
\hline
Embedding Method & Test Acc. & Test loss & Time (s)  \\ \hline
\textit{Agg1R} & 81.2\% & 0.398 & 0.402 \\ \hline
\textit{Agg1}  & 81.5\% & 0.135 & 0.412 \\ \hline
\textbf{\textit{Agg2}}  & \textbf{82.3}\% & \textbf{0.144} & \textbf{0.401} \\ \hline
\textit{Agg2R} &79.7\% &  0.41 & 0.426 \\ \hline
\textit{EVC1}  & 80.3\% & 0.156 & 0.431 \\ \hline
\textit{EVC2}  & 80.4\% & 0.187 & 0.43 \\ \hline
\textit{EVC2R} & 78.8\% & 0.409 & 0.455 \\ \hline
\textit{EVC1R} & 79.5\% & 0.394 & 0.433 \\ \hline
\end{tabular}
\end{table}

\begin{table}[!h]
\caption{Performance results of \textit{flexgrid2vec} with patch size of $64 \times 64$.}\label{64_results}
\small
\center
\begin{tabular}{|l|l|l|l|}
\hline
Embedding Method & Test Acc. & Test loss & Time (s)  \\ \hline
\textit{Agg1R} & 90.1\% & 0.105 & 0.039 \\ \hline
\textit{Agg2R} & 90\% & 0.227 & 0.053 \\ \hline
\textbf{\textit{EVC1R}} & \textbf{90.9\%} & \textbf{0.229} & \textbf{0.041} \\ \hline
\textit{EVC2R} & 89.4\% & 0.263 & 0.057  \\ \hline
\end{tabular}
\end{table}

\subsubsection{Comparison with other embedding methods}
For benchmarking the embedding component of \textit{flexgrid2vec}, we utilised the state-of-the-art node2vec \cite{grover2016node2vec} algorithm to compute the node embeddings. node2vec is useful to produce flexible node neighbourhoods. The neighbourhood design is biased to random walks. This process enables the algorithm to learn a low-dimensional feature space based on the skip-gram that is utilised in the deep-walk methodology \cite{perozzi2014deepwalk}. The aim now is to maximise node neighbourhood log-probability, as in the following objective function:
\begin{equation}\label{node2vec1}
  \max_{f} \sum_{v \in V} \log \Pr (N(v)|f(v))
\end{equation}
The edge weights are used to bias the random walks to control the next node that will be considered in the neighbourhood. In addition, we also implemented a layer message aggregation network based on GCN \cite{kipf2017semi}. The feature vector is computed based on the grid-graph representations. 

Tables \ref{nodebinary} and \ref{nodemc} show that \textit{flexgrid2vec} outperforms both models with high accuracy margins in both binary and multi-class image classification. \textit{flexgrid2vec} has $25$ and $10$ percent more test accuracy in comparison to using GCN and node2vec.

\begin{table}[!h]
\begin{center}
\caption{Benchmarking the node embedding models for binary classification}\label{nodebinary}
\small
\centering
\begin{tabular}{|p{3cm}|l|l|}
\hline
Model & Test Accuracy & Test Loss \\ \hline
flexgrid2vec (node2vec) 
& 73.0\% & 0.122 \\ \hline
flexgrid2vec (GCN) 
& 89.2\% & 0.271 \\ \hline
\textbf{flexgrid2vec} &
\textbf{98.8\%} & 
\textbf{0.049} \\ \hline
\end{tabular}
\end{center}
\end{table}

\begin{table}[!h]
\begin{center}
\caption{Benchmarking with the node embedding models for Multi-class classification}\label{nodemc}
\small
\begin{tabular}{|p{4.5cm}|l|l|}
\hline
Model & Test Accuracy & Test Loss \\ \hline
flexgrid2vec (node2vec) \cite{grover2016node2vec} & 36.4\% & 2.043 \\ \hline
flexgrid2vec (GCN) \cite{kipf2017semi} & 70.3\% & 1.020 \\ \hline
\textbf{flexgrid2vec}&
\textbf{89.74\%} & 
\textbf{0.11} \\ \hline
\end{tabular}
\end{center}
\end{table}

\subsection{Visual Feature Representation}
\begin{figure}[!ht]
\begin{center}
  \includegraphics[angle=0, origin=c, width=0.99\linewidth]{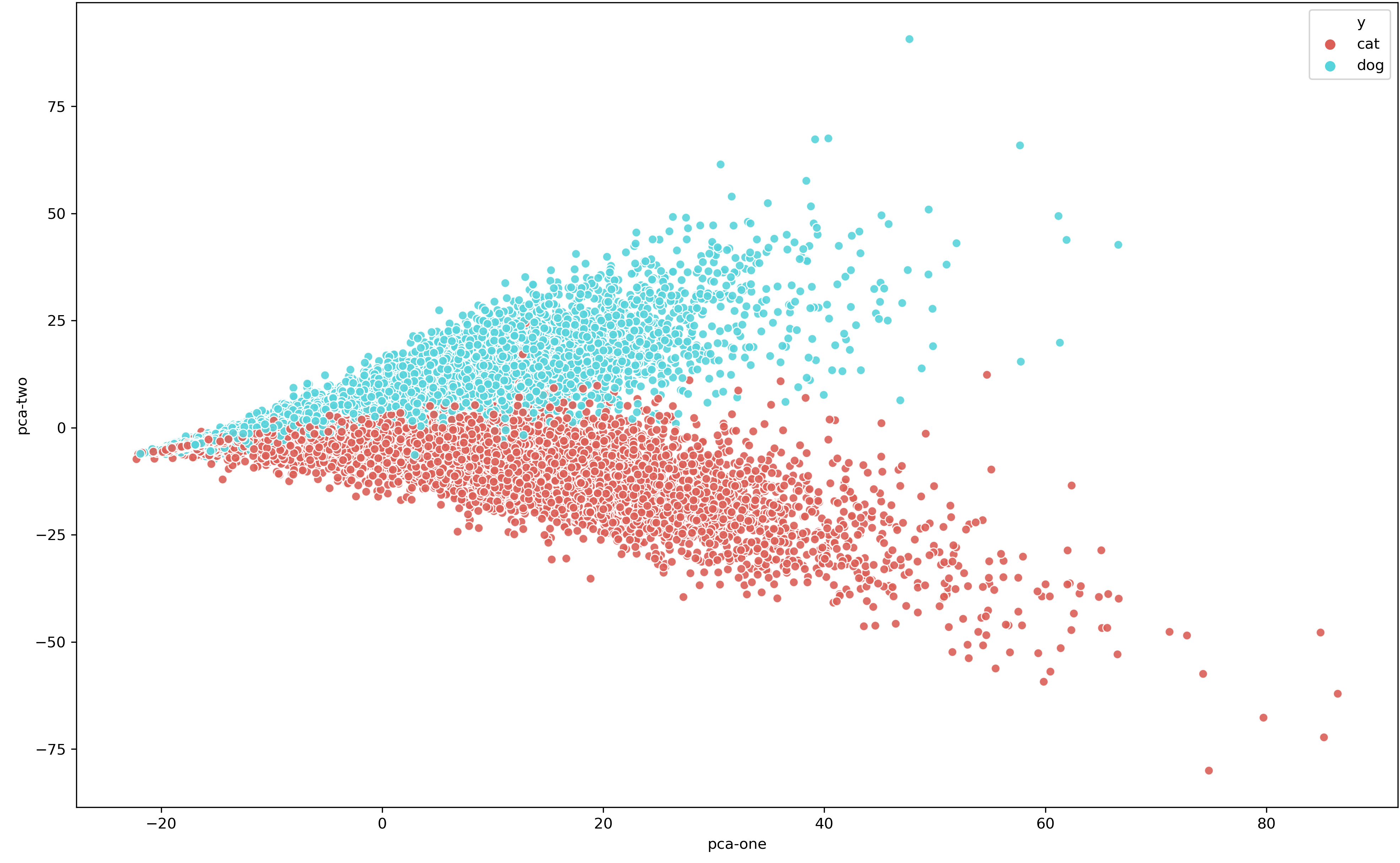}
\end{center}
  \caption{PCA two-component visualisation of ASIRRA dataset \textit{flexgrid2vec} vectors.}
\label{pca_cd}
\end{figure}

\begin{figure}[!ht]
\begin{center}
  \includegraphics[angle=0, origin=c, width=0.99\linewidth]{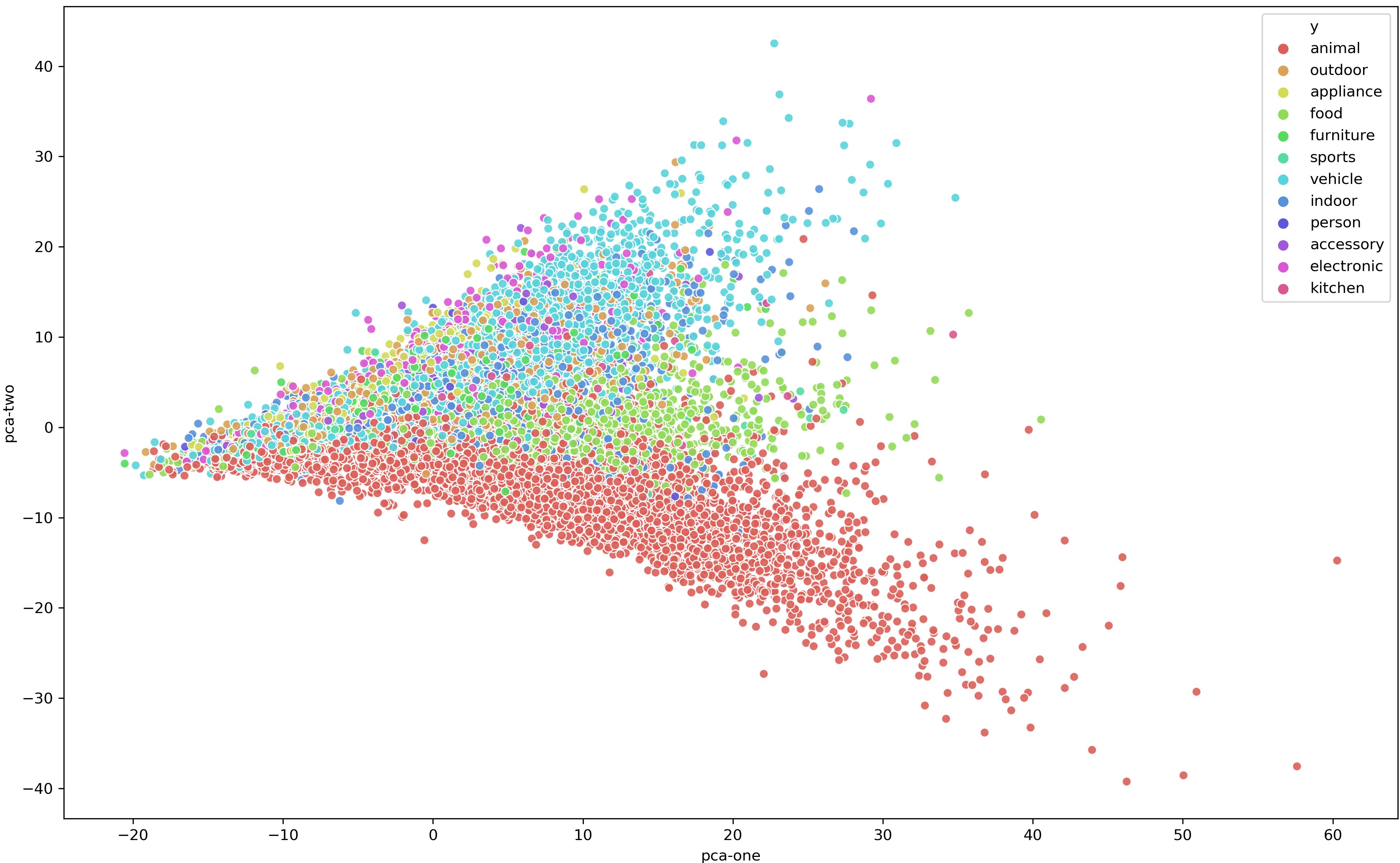}
\end{center}
  \caption{PCA two-component visualisation of COCO dataset \textit{flexgrid2vec} vectors.}
\label{pca_coco}
\end{figure}

\textit{flexgrid2vec} is a visual representation algorithm that represents an image as a feature vector. Fig. \ref{pca_cd} and Fig. \ref{pca_coco} show 2D visualisations of \textit{flexgrid2vec} representations of two-class (ASIRRA) and 12-class (COCO) datasets. \textit{flexgrid2vec} produces a $1 \times 512$ feature vector. We utilised principal component analysis (PCA) to reduce the dimensions for visualisation purposes. Fig. \ref{pca_cd} highlights the superiority of \textit{flexgrid2vec}’s performance. The given ASIRRA image dataset is successfully represented in a significant way. This high performance is supported by the classification accuracy. Fig. \ref{pca_coco} shows the \textit{flexgrid2vec} performance on the COCO image dataset. The given dataset is highly imbalanced, yet \textit{flexgrid2vec} still produces discriminative features, as illustrated in Fig. \ref{pca_coco}.


\section{Conclusion}\label{conclusion}
We propose a novel GNN-based approach to learn visual features, called \textit{flexgrid2vec}. It learns feature representation through constructing grid-graphs that are flexible to the input image. The experiment results show the superiority of the proposed \textit{flexgrid2vec} in both binary and multi-class image classification using reduced feature vectors. \textit{flexgrid2vec} learns image features of only the most important patches.  In the future, we will investigate the usage of \textit{flexgrid2vec} for few-shot learning, visual segmentation and object detection. We will also research the utilisation of other loss functions, such as contrastive loss and data augmentation techniques.

\section*{Acknowledgments}
Ali Hamdi is supported by RMIT Research Stipend Scholarship. This research is partially supported by Australian Research Council (ARC) Discovery Project \textit{DP190101485}.

\bibliographystyle{elsarticle-num}
\bibliography{bib}

\end{document}